\documentclass[10pt,twocolumn,letterpaper]{article}

\usepackage{cvpr}
\usepackage{times}
\usepackage{epsfig}
\usepackage{graphicx}
\usepackage{amsmath}
\usepackage{amssymb}

\def\etal{\emph{et al}.}
\def\ie{\emph{i.e}.} 

\usepackage[svgnames,table]{xcolor} 
\usepackage{multirow}
\usepackage{soul}
\usepackage{amssymb}  

\definecolor{rgb1}{RGB}{214,  38, 40}   
\definecolor{rgb2}{RGB}{43, 160, 4}     
\definecolor{rgb3}{RGB}{158, 216, 229}  
\definecolor{rgb4}{RGB}{114, 158, 206}  
\definecolor{rgb5}{RGB}{204, 204, 91}   
\definecolor{rgb6}{RGB}{255, 186, 119}  
\definecolor{rgb7}{RGB}{147, 102, 188}  
\definecolor{rgb8}{RGB}{30, 119, 181}   
\definecolor{rgb9}{RGB}{160, 188, 33}   
\definecolor{rgb10}{RGB}{255, 127, 12}  
\definecolor{rgb11}{RGB}{196, 175, 214} 

\cvprfinalcopy 

\begin{document}

\title{3D Gated Recurrent Fusion for Semantic Scene Completion}


\author{Yu Liu$^{1^\dag}$, Jie Li$^{2^\dag}$, Qingsen Yan$^{3}$,
Xia Yuan$^{2}$, Chunxia Zhao$^{2}$, Ian Reid$^{1}$ and Cesar Cadena$^{4}$
\thanks{
\noindent
\textit{Yu Liu and Jie Li contributed equally to this work.}
\newline
$^{1}$Y. Liu and I. Reid 
are with School of Computer Science, The University of Adelaide,
5005, North Terrace, SA 
{\tt\small yu.liu04@adelaide.edu.au}
\newline
$^{2}$J. Li, X. Yuan and C. Zhao 
are with School of Computer Science and Engineering, Nanjing University of Science and Technology, Nanjing, 210094, China {\tt\small jieli\_cn@163.com}
\newline
$^{3}$ Q. Yan is with School of Computer Science and Engineering, Northwestern Polytechnical University, Xi'an, 710072, China {\tt\small yqs@mail.nwpu.edu.cn}
\newline
$^{4}$C. Cadena 
is with Autonomous Systems Lab, ETH Zurich, 
Leonhardstrasse 21, 8092, Zurich 
{\tt\small cesarc@ethz.ch}
}
}

\maketitle

\begin{abstract}
This paper tackles the problem of data fusion in the semantic scene completion (SSC) task, which can simultaneously deal with semantic labeling and scene completion. RGB images contain texture details of the object(s) which are vital for semantic scene understanding. Meanwhile, depth images capture geometric clues of high relevance for shape completion. Using both RGB and depth images can further boost the accuracy of SSC over employing one modality in isolation. We propose a 3D gated recurrent fusion network (GRFNet), which learns to adaptively select and fuse the relevant information from depth and RGB by making use of the gate and memory modules.
Based on the single-stage fusion, we further propose a multi-stage fusion strategy, which could model the correlations among different stages within the network. Extensive experiments on two benchmark datasets demonstrate the superior performance and the effectiveness of the proposed GRFNet for data fusion in SSC. Code will be made available.

\end{abstract}

\section{Introduction}

Understanding the surroundings is a fundamental capability for many real-world applications such as augmented reality~\cite{chen2017semantic}, robot grasping~\cite{varley2017shape}, or autonomous navigation~\cite{doan2019scalable}.
Different abstractions are possible, and even complementary. 
Semantic labeling of the scene allows for a high level reasoning, while 3D geometry completion enables basic spatial capabilities.
Semantic scene completion (SSC) aims at solving both simultaneously.

An RGB-D sensor allows acquiring depth information from the scene along side the RGB image. 
On the one hand, RGB image contains rich details about the color and texture, which are the primary cues for the semantic scene understanding. 
On the other hand, depth carries more clues about the object geometry and distance information, which are much reliable in reflecting the position, shape, and occlusion relationship between objects within the scene.
%
%
Many vision applications have already benefit from using both modalities in their tasks, such as object detection~\cite{gupta2014learning,chen2017multi}, video segmentation~\cite{fu2017object,sultana2018unsupervised,emre2017semantic}, action recognition~\cite{ijjina2017human,hu2018deep, zhang2018rgb}, 
or visual SLAM~\cite{kerl2013dense,whelan2013robust,lu2015robust}.
Recent studies~\cite{Garbade2018_twoStream,li2019rgbd} in SSC also demonstrate that employing both, RGB image and depth, can outperform using only one modality~\cite{song2017_SSCNet}.

However, fusing the information from RGB and depth is still an unsolved problem, and becomes an
obstacle which hinders the performance of SSC.
Albeit some recent works conduct data fusion between RGB and depth, they usually employ some, ``manually'' set, basic operation to fuse the data. 
Those includes \textit{sum fusion}~\cite{li2019rgbd,hazirbas2016fusenet}, \textit{max fusion}~\cite{kang2014convolutional}, \textit{concatenate fusion}~\cite{couprie2013indoor,guo2018semantic}, 
\textit{transform fusion}~\cite{wang2016learning} and \textit{bilinear fusion}~\cite{lin2015bilinear}. Nevertheless, RGB and depth data are not equivalent quantities, while still providing complementary yet redundant information. Therefore, we propose to extract the information in a selective manner from both modalities, and fuse them accordingly with respect to the specific task.

We present the Gated Recurrent Fusion (GRF) block, which can provide adaptive selection and aggregation of RGB and depth information. On the one hand, the \emph{gate} component in the GRF fusion block selects, in an adaptive manner, various positions of different importance in aligned RGB-D frames regarding to the contribution from both modalities.
The \emph{gate} effectively selects valid information while filters out the irrelevant one.
On the other hand, the
\emph{memory} component in the GRF fusion block effectively preserves the complementary information, which can compensate the missing or ambiguous details of the data obtained from different modalities. 

Furthermore, the GRF fusion block offers the flexibility to be cascaded to a multi-stage configuration that combines high-level and low-level features.
GRF fusion block is extended from the Gated Recurrent Unit (GRU)~\cite{cho2014learning}. 
Based on the GRF fusion block, we build the GRFNet for the semantic scene completion, and provide single- and multi-stage fusion versions of GRFNet. 
In the single-stage fusion version, depth and RGB features are fed into the same GRF fusion block individually. 
In the multi-stage fusion version, depth and RGB features of different stages form an interleaved sequence and are input into the same GRF fusion block consecutively. 

The multi-stage version takes advantage of both low-level and high-level features, and achieves better performance than the single-stage version.

In summary, the contributions of this work are mainly two-fold:
\vspace{-0.1cm}
\begin{itemize}
    \item
    An end-to-end 3D-GRF based network, GRFNet, is presented for fusing RGB and depth information in the SSC task,
    through employing \emph{gate} and \emph{memory} components, the selection and fusion between two modalities can be conducted effectively. 
    To the best of our knowledge, this is the first time that gated recurrent network is employed for data fusion in the SSC task.
    \vspace{-0.1cm}
    
    \item 
    Within the framework of GRFNet, single-stage and multi-stage fusion strategies are proposed. While outperforming existing fusing strategies in the SSC task already with the single stage, the multi-stage fusion proves to give the best results.
\end{itemize}
\vspace{-0.1cm}
Extensive experiments demonstrate that the proposed GRFNet achieves superior performance on NYU~\cite{silberman2012indoor} and NYUCAD~\cite{firman2016NYUCAD} datasets.

\begin{figure}[t]
\begin{center}   
{
\includegraphics[width=0.8\linewidth]{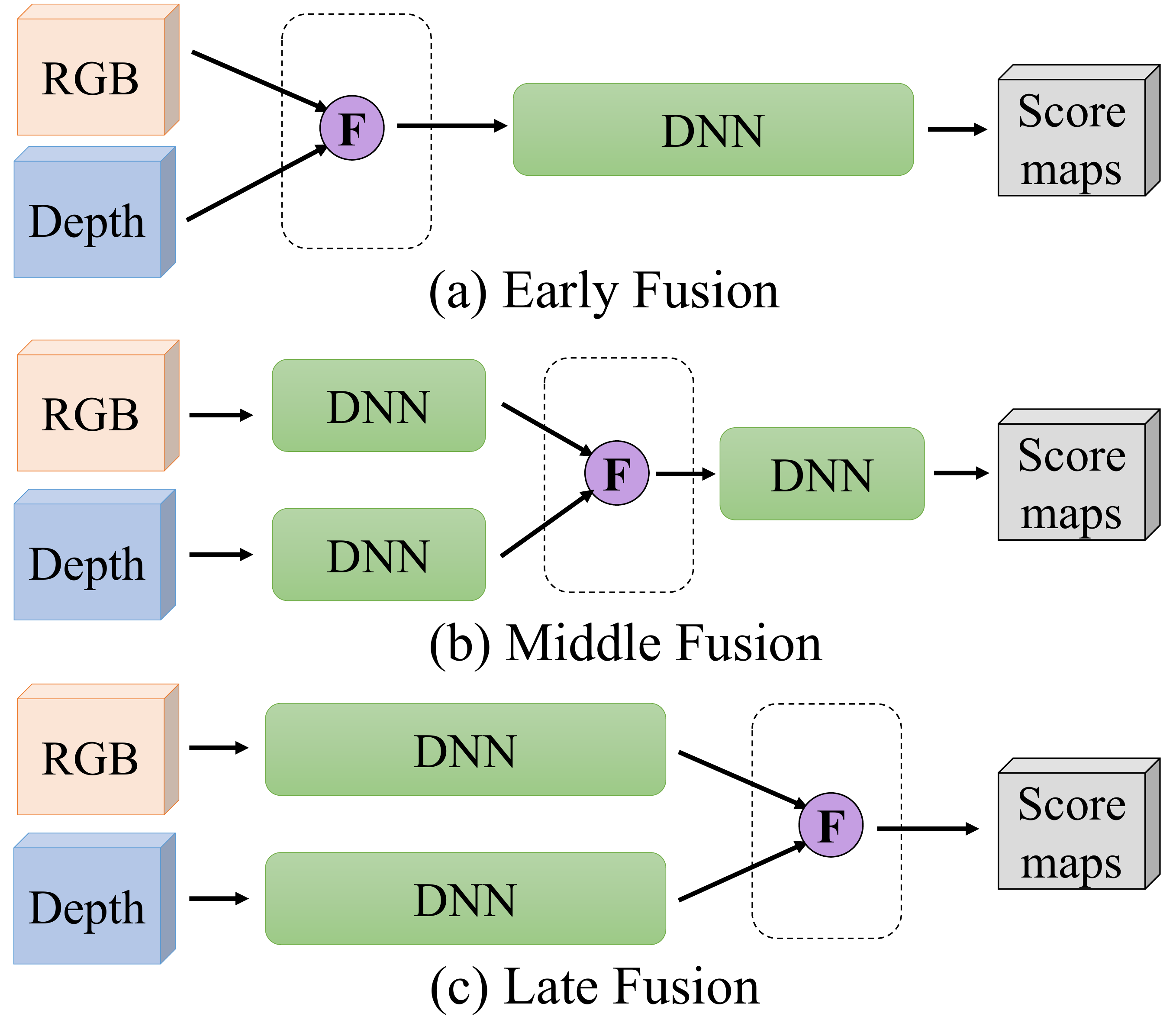}
}
\caption{Fusion at different stages. 
Early fusion contains more low-level features, while the input of the late fusion contains more abstract high-level features.
}
\vspace{-0.7cm}
\label{fig:fusion-when}
\end{center}
\end{figure}
\section{Related Work}
In this section, we briefly look through the deep learning based methods for SSC, with emphasis on the discussion of the existing multi-modality fusion strategies.

\vspace{-0.1cm}
\subsection{Semantic Scene Completion}

The goal of SSC is to produce a complete 3D voxel representation for a scene from a single-view input.
Specifically, Song \etal~(\cite{song2017_SSCNet}) propose an end-to-end 3D convolutional network (SSCNet) which is based on the single-view depth as input that can simultaneously predict the results of scene completion and semantic labeling.
SSCNet has high computational costs due to the adoption of 3D convolutions. 
Zhang \etal~(\cite{zhang2018efficient}) introduce spatial group convolution (SGC) into SSC for accelerating the computation of 3D dense prediction task. 
Meanwhile, Han \etal~(\cite{han2017high}) employ the long short-term memory (LSTM) to recover missing parts of 3D shapes.
Dai \etal~(\cite{dai2018scancomplete}) use a coarse-to-fine strategy to handle large scenes with varying spatial extent.
Although the depth-based approach has made significant progress, the absence of texture details prevents improving SSC.  

In order to incorporating the color information, TS3D~\cite{Garbade2018_twoStream} introduces the RGB image into SSC and uses a 2D network to acquire semantic segmentation results.
Semantic outputs of the RGB stream are concatenated with inputs of the depth stream to obtain the completed 3D scene. DDR-SSC~\cite{li2019rgbd} uses two parallel feature extraction branches with the same structure to obtain information from RGB and depth simultaneously.
A multi-stage structure with element-wise addition is employed to perform feature fusion.
Thanks to the semantic information provided by RGB, the semantic labeling accuracy of both TS3D and DDR-SSC has significantly been improved compared to SSCNet. However, none of these methods takes into account the selective fusion of multi-modal information, which limit those algorithms to achieve better performance.


\subsection{Fusion Schemes}
\vspace{-0.1cm}
The RGB-D information fusion is important to many vision applications.
In general, the fusion scheme can be divided into three categories, \textit{e.g.} early fusion~\cite{couprie2013indoor}, middle fusion~\cite{ren2012rgb} and late fusion~\cite{yue2015beyond,simonyan2014two}, as shown in Figure~\ref{fig:fusion-when}.
According to the stages of fusion, these schemes can also be divided into single-stage fusion and multi-stage fusion \cite{hazirbas2016fusenet,park2017rdfnet}.

\begin{figure}[t]
\begin{center}   
{
\includegraphics[width=0.88\linewidth]{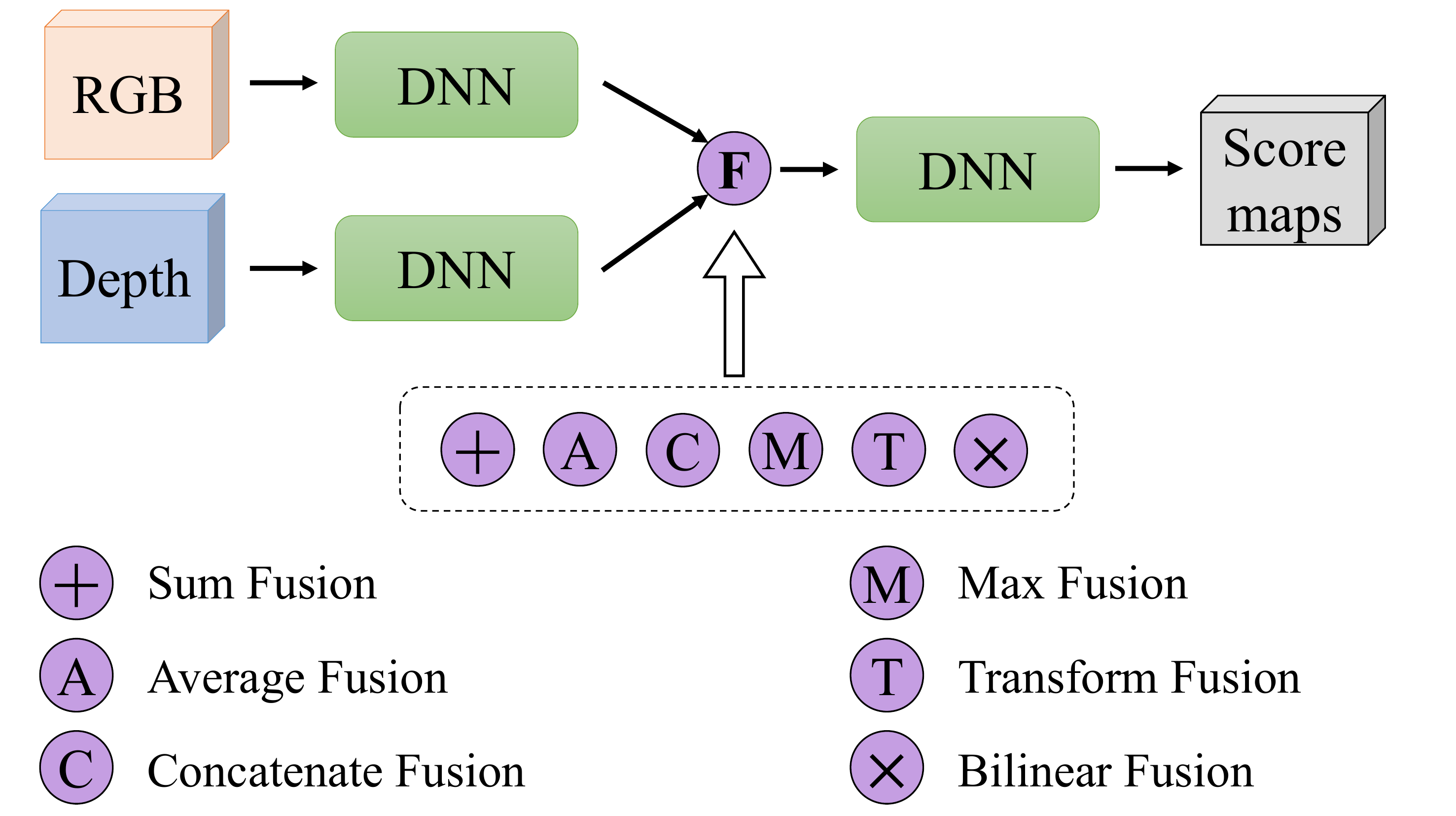}
}
\caption{Several typical single-stage fusion methods. }
\vspace{-0.5cm}
\label{fig:fusion-single-stage}
\end{center}
\end{figure}

\begin{figure*}[t]
\begin{center}
{
\includegraphics[width=0.99\linewidth]{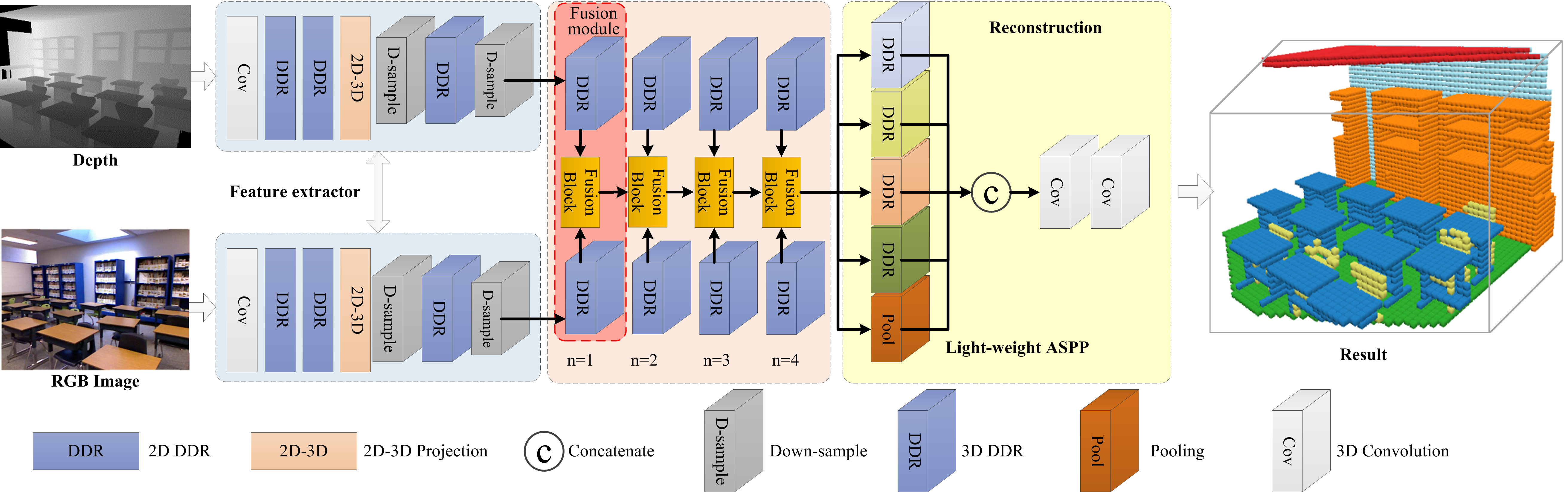}
}
\caption{The network architecture of GRFNet is extended from Dimensional Decomposition Residual (DDR) network~\cite{li2019rgbd}. 
GRFNet has two feature extractors to capture the features from depth and RGB images respectively.
The feature extractor contains a projection layer to map 2D feature to 3D space.
The GRF fusion block (denoted by yellow boxes in the middle) replaces the original fusion unit to take full advantage of the multi-modal information. With two DDR plus their corresponding GRF fusion block to form a fusion module, also named single-stage fusion module (denoted by the red box in one column). GRFNet is composed of a multi-stage ( 4-stage here) fusion module. Then we use light-weight ASPP to obtain multiple receptive fields information.
Different colors of the DDR block denote various receptive fields.
Then the network uses two 3D convolutions to predict occupancies and object labels simultaneously.
}
\label{fig:network_structure}
\end{center}
\end{figure*}

\noindent
\textbf{Single-Stage Fusion} There are several general patterns for single-stage fusion as shown in Figure~\ref{fig:fusion-single-stage}. Specifically, Sum fusion~\cite{li2019rgbd,hazirbas2016fusenet} computes the sum of the two feature maps at the same spatial locations.
Average fusion is essentially a weighted sum fusion with equal weights.
Max fusion~\cite{kang2014convolutional} takes the feature with the maximal value from multiple feature maps.
Concatenate fusion stacks the features with channels~\cite{couprie2013indoor,guo2018semantic}.
Wang \emph{et al.}~(\cite{wang2016learning}) propose an encoder-decoder architecture which exchanges the information of multi-modal data in the latent space.
Bilinear fusion~\cite{lin2015bilinear} computes an outer matrix product of the two features at each pixel location.

There are a few methods that consider the complementary and selectivity of data fusion.
Specifically, Li \etal~(\cite{li2016lstm-cf}) develop a novel LSTM model to fuse scene contexts adaptively.
Cheng \etal~(\cite{cheng2017locality}) use the concatenated feature maps of RGB and depth to learn an array $G$ to weight the contribution of one input modality and $1-G$ to weight the other input modality.
Wang \etal~(\cite{wang2019adaptive}) use the same strategy as~\cite{cheng2017locality} to fuse the feature maps from RGB and Depth in saliency detection.
However, Li \etal~(\cite{li2016lstm-cf}) only consider the complementarity of information but ignore the selectivity of the data, the other two methods only consider the selectivity of information and cannot guarantee the complementarity of information.
Moreover, these methods are single-stage fusion and lack of scalability.

\noindent
\textbf{Multi-stage Fusion} 
According to the way for aggregating multi-modal information, this paper divides multi-stage fusion algorithms into merge fusion, cross fusion, and external fusion.
Hazirbas \etal~(\cite{hazirbas2016fusenet}) adopt the merge fusion structure to fuse the two branches of features extracted from RGB and depth images.
The feature maps from depth are fused into the RGB branch by stages with an element-wise summation.
Wang \etal~(\cite{wang2016learning}) use cross fusion to merge the common features of RGB and depth, and keep the modality specific features separated from each other.
Both Park \etal~(\cite{park2017rdfnet}) and Li \etal~(\cite{li2019rgbd}) use an external fusion mechanism. Specifically, Li \etal~(\cite{li2019rgbd}) capture features of RGB and depth image at different levels, these features at each level are fused separately and then assembled all at once before the reconstruction part. Park \etal~(\cite{park2017rdfnet}) propose RDFNet to fuse multi-modal features separately by multiple fusion blocks,
and refine the fused features one by one through a set of refine blocks. In RDFNet, each fusion introduces an additional fusion block with a new set of extra parameters.
The artificially designed fusion blocks are complex and require multiple parameters that are not easy to migrate to other applications.
These multi-stage fusion methods use high-level and low-level features achieving high accuracy. However, each fusion block within the multi-stage mostly adopts concatenation or summation, ignoring the adaptive selection of the multi-modal data.

On the contrary, our proposed GRF fusion block extends the standard gated recurrent unit (GRU), where the gate and the memory structures can adaptively select and preserve valid information. Besides, GRFNet adopts the form of a recurrent network. When performing multi-stage fusion, GRF modules exploit parameter sharing.
And experiments show that both of the proposed single- and multi-stage GRFNets achieve better accuracy than previous methods.



\section{Methodology}
\vspace{-0.1cm}
\subsection{Overview}
\vspace{-0.1cm}

Our proposal, GRFNet, extends the network architecture of DDR-SSC~\cite{li2019rgbd}, and focuses on improving the fusion strategy.
We subtly adopt the gate structure and memory mechanism in GRU unit to form a multi-modal feature GRF fusion block with the power of autonomous selectivity and adaptive memory preservation.
Moreover, taking advantage of its recurrent nature, we further propose a multi-stage fusion strategy to utilize both low-level and high-level features with introducing insignificant parameters. 

In the feature extractor part, the network uses dimensional decomposition residual (DDR) blocks to extract the local textures and the geometry information. A projection layer is employed to connect the 2D and 3D parts. The multi-stage fusion module consists of four single-stage fusion modules that can effectively combines the RGB features and depth features.
The fused features are fed into the subsequent light-weight atrous spatial pyramid pooling (LW-ASPP~\cite{li2019rgbd}).
After that, another two point-wise convolutional layers are used to predict the semantic labels for each voxel in the 3D volume. 

The network maps each voxel to one of the labels $C = {c_0, c_1,\cdots c_{N}}$, where $N$ is the number of semantic classes, and $c_0$ represents the empty voxel.

\subsection{Gated Recurrent Unit}
Gated recurrent unit (GRU)~\cite{cho2014learning} is a popular model in recurrent neural networks (RNN) and has an outstanding performance in many natural language processing (NLP) tasks~\cite{chorowski2015attention,kim2016character}.
A GRU has two gate structures, a memory structure, and can be reused recurrently.
However, few researchers have explored the power of GRU in the field of 3D vision, especially for feature fusion.
We find that GRU highly aligns with our requirements for an effective multi-modal fusion strategy in SSC. 

The gate structure in GRU enables the selective fusion of multi-modal features.
The memory structure ensures that valid information can be retained for future fusion purpose.
The characteristics of its recurrent network enable GRU to be reused in the multi-stage fusion while sharing the same set of parameters.
%
Compared to GRU, the structure of Long short-term memory (LSTM)~\cite{hochreiter1997long} is more complicated and has an extra forget gate with more parameters.
ConvGRU~\cite{ballas2015delving} is a convolutional version of GRU.  We extend ConvGRU to 3D convolutional in our GRFNet, and modify it to fit the feature fusion in SSC task.

\subsection{Gated Recurrent Fusion Block of RGB-D Features}

\begin{figure}[t]
\begin{center}   
{
\includegraphics[width=0.99\linewidth]{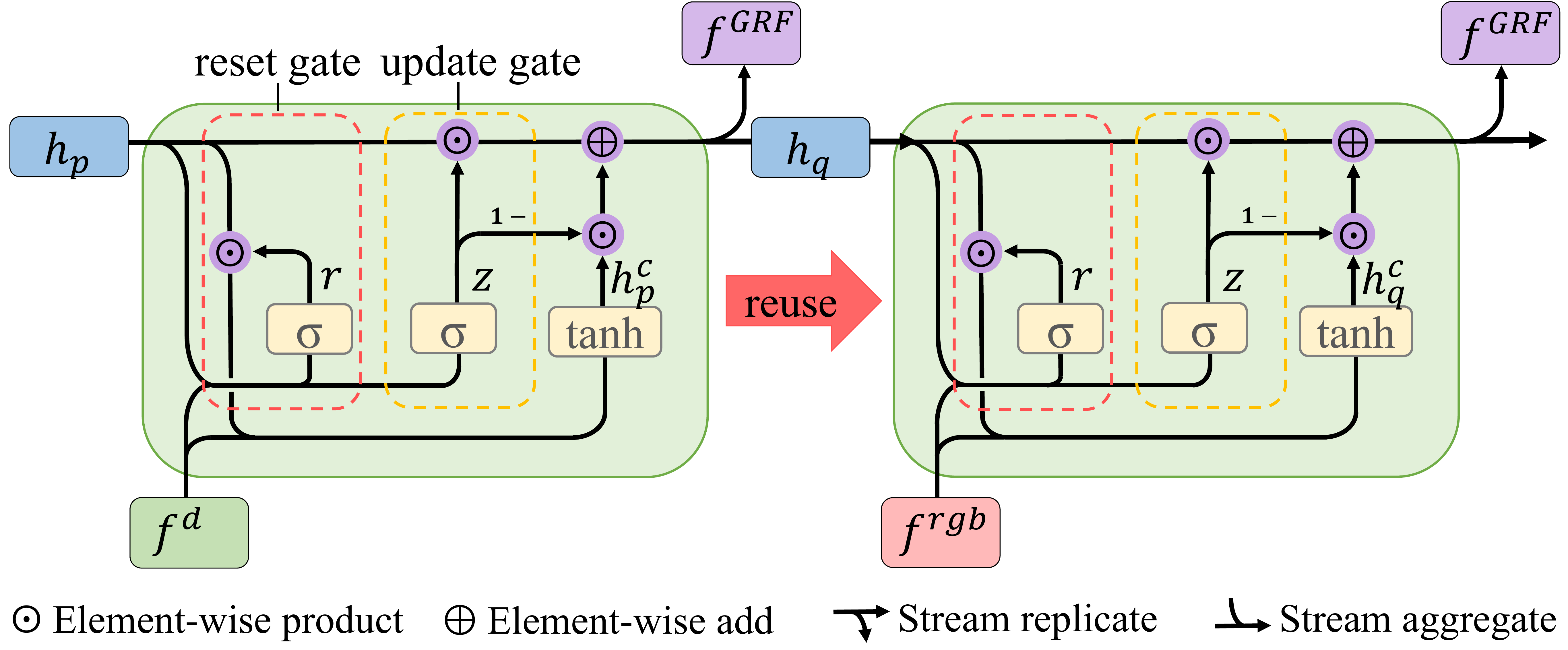}
}
\caption{GRF fusion block. At step $p$, the input of GRF fusion block is one of the features from depth or RGB, and in the next step, input is the other. Both GRF fusion blocks share the same set of parameters.}
\vspace{-0.5cm}
\label{fig:GRUFusion}
\end{center}
\end{figure}

As shown in Figure~\ref{fig:GRUFusion}, at fist step (left), gated recurrent fusion (GRF) block takes one of the RGB-D features as input.
The outputs of this step will be used as the hidden state.
Then, in the second step (right), GRF fusion block takes the features of another modality as input. These two steps reuse the GRF fusion block and share the same set of parameters.
Next, we will use the first step with input $f^d$ as an example to explain in detail the principle and work flow of the GRF fusion block. 

In contrast to the commonly used GRU for encoding information in a temporal way, the way we use GRF to fuse RGB+Depth features is somehow different. Specifically, the GRU handles the fusion of RGB and depth information in a `modality', 
rather than a `sequential', way. 
And for the multi-stage fusion process, similar as other deep neural networks, it is more like the low-level multi-modal feature to guide the following high-level multi-modal feature to be merged.

\begin{table*}[t]
\begin{center}
\scalebox{0.8}
{
\begin{tabular}{l |c c c|c c c c c c c c c c c c} 
\hline
  & \multicolumn{3}{c|}{scene completion} & \multicolumn{12}{c}{semantic scene completion} \\ \hline
Method  & prec. & recall & IoU & \cellcolor{rgb1}ceil. & \cellcolor{rgb2}floor & \cellcolor{rgb3}wall & \cellcolor{rgb4}win. & \cellcolor{rgb5}chair & \cellcolor{rgb6}bed & \cellcolor{rgb7}sofa & \cellcolor{rgb8}table & \cellcolor{rgb9}tvs & \cellcolor{rgb10}furn. & \cellcolor{rgb11}objs. & avg. \\ 
\hline
Lin \etal~(\cite{lin2013holistic})
& 58.5 & 49.9 & 36.4 &  0.0 & 11.7 & 13.3 & 14.1 &  9.4 & 29.0 & 24.0 &  6.0 &  7.0 & 16.2 &  1.1 & 12.0\\
Geiger \etal~(\cite{geiger2015joint}) & 65.7 & 58.0 & 44.4 & 10.2 & 62.5 & 19.1 &  5.8 &  8.5 & 40.6 & 27.7 &  7.0 &  6.0 & 22.6 &  5.9 & 19.6\\ 
SSCNet~\cite{song2017_SSCNet} & 57.0 & {\bfseries 94.5} & 55.1 & 15.1 & {\bfseries 94.7} & 24.4 &  0.0 & 12.6 & 32.1 & 35.0 & 13.0 &  7.8 & 27.1 & 10.1 & 24.7\\
EsscNet~\cite{zhang2018efficient}   & {\bfseries 71.9} & 71.9 & 56.2 & 17.5 & 75.4 & 25.8 &  6.7 & 15.3 & {\bfseries 53.8} & 42.4 & 11.2 &    0 & 33.4 & 11.8 & 26.7\\ 

DDR-SSC~\cite{li2019rgbd}  & 71.5  & 80.8 & 61.0 & 21.1 & 92.2 & {\bfseries 33.5} & 6.8 & 14.8 & 48.3 & 42.3 &  13.2 & {\bfseries 13.9} & 35.3 & 13.2 & 30.4 \\ 

GRFNet & 68.4 & 85.4 & {\bfseries 61.2}  &  {\bfseries 24.0} & 91.7 & 33.3 & {\bfseries 19.0} & {\bfseries 18.1} & 51.9 & {\bfseries 45.5} & {\bfseries 13.4} &  13.3 & {\bfseries 37.3} & {\bfseries 15.0}  & {\bfseries 32.9}  \\
\hline
\end{tabular}
}
\caption{Results on the NYU dataset~\cite{silberman2012indoor}. Bold numbers represent the best scores.}
\vspace{-0.3cm}
\label{tab:NYU}
\end{center}
\end{table*}

\begin{table*}[t]
\begin{center}
\scalebox{0.8}
{
\begin{tabular} {l |c c c|c c c c c c c c c c c|c} \hline
 &  \multicolumn{3}{c|}{scene completion} & \multicolumn{12}{c}{semantic scene completion} \\ 
\hline
Method  & prec. & recall & IoU & \cellcolor{rgb1}ceil. & \cellcolor{rgb2}floor & \cellcolor{rgb3}wall & \cellcolor{rgb4}win. & \cellcolor{rgb5}chair & \cellcolor{rgb6}bed & \cellcolor{rgb7}sofa & \cellcolor{rgb8}table & \cellcolor{rgb9}tvs & \cellcolor{rgb10}furn. & \cellcolor{rgb11}objs. & avg. \\ 
\hline
Zheng \etal(~\cite{zheng2013beyond}) 	& 60.1 & 46.7 & 34.6 & - & - & - & - & - & - & - & - & - & - & - & - \\ 
Firman \etal(~\cite{firman2016NYUCAD}) 	& 66.5 & 69.7 & 50.8 & - & - & - & - & - & - & - & - & - & - & - & - \\ 
SSCNet~\cite{song2017_SSCNet}   & 75.4 & {\bfseries 96.3} & 73.2 & 32.5 & 92.6 & 40.2 &  8.9 & 33.9 & 57.0 & {\bfseries 59.5} & 28.3 &  8.1 &  44.8 & 25.1 & 40.0\\ 
TS3D~\cite{Garbade2018_twoStream}      	& 80.2 & 91.0 & 74.2 & 33.8 & {\bfseries 92.9} & 46.8 & {\bfseries 27.0} & 27.9 & {\bfseries 61.6} & 51.6 & 27.6 & {\bfseries 26.9} & 44.5 & 22.0 & 42.1\\ 
DDR-SSC~\cite{li2019rgbd}   & {\bfseries 88.7} & 88.5 & 79.4 & {\bfseries 54.1} & 91.5 & 56.4 & 14.9 & 37.0 & 55.7 & 51.0 &  28.8 & 9.2 & 44.1 &  27.8 & 42.8 \\

GRFNet        	& 87.2 & 91.0 & {\bfseries 80.1} & 50.3& 91.8& {\bfseries 58.1 } & 18.4 & {\bfseries 42.7} & 60.6 & 52.8 & {\bfseries 34.6 }&  11.5 & {\bfseries 46.6} & {\bfseries 30.8} & {\bfseries 45.3} \\
\hline
\end{tabular}
}  
\caption{Results on the NYUCAD dataset~\cite{zheng2013beyond}. Bold numbers represent the best scores.
}
\label{tab:NYUCAD}
\end{center}
\end{table*}

\noindent
\textbf{Hidden State}
The hidden state ${h}_{p}$, along with the current input $f$ to control the reset and update gates.
The output of the previous stage will be used as the hidden state of current stage.
At the first step of fusion between depth feature $f^{d}$ and RGB feature $f^{rgb}$, that is $p=1$, we use the sum fusion of two modal features to initialize the hidden state, as ${h}_{0} = f^{d} + f^{rgb}$.

\noindent
\textbf{Reset Gate}
At step $p$, the hidden state ${ h }_{ p }$ and the current input $f^{d}$ together to decide the status of the reset gate $r$ by
\begin{equation}\label{Eq:convGRU-r}
{ r }=\sigma \left( { W }_{ r }\left({ f }^{ d }, { h }_{ p } \right) \right)
\end{equation}

The two feature stream ${ f }^{ d }$ and ${ h }_{ p }$ are concatenated and fed into a convolution operation. ${ W }_{ r }$ represents the corresponding weights in the convolution.
The sigmoid function $\sigma$ converts each value in the feature tensor into the range of $(0,1)$ and acts as a gate signal.

\noindent
\textbf{Update Gate}
The update gate $z$ is also decided by the hidden state ${ h }_{ p }$ and input features ${ f }^{ d }$.
Through another convolution operation with weight ${ W }_{ z }$ and the sigmoid function $\sigma$, we get $z$ as,
\begin{equation}\label{Eq:convGRU-z}
{ z }=\sigma \left( { W }_{ z }\left( { f }^{ d }, { h }_{ p } \right)\right)
\end{equation}

Theoretically, reset and update gates essentially learn a set of weights that control the amount of information that is retained or discarded, 
experimental studies in section~\ref{sec:experiments} show the effectiveness of the reset and update gates.

\noindent
\textbf{Adaptive Memory}
Through the element-wise product $\odot$, the reset gate $r$ determines how much information in the past needs to be ``memorized''.
\begin{equation}\label{Eq:convGRU=h_t-1}
{h'}_{ p }={ r }\odot { h }_{ p }
\end{equation}
when the reset gate $r$ is close to 1, the ``memorized'' information ${h'}_{ p }$ will be kept and then passed to the current fusion operation with current input feature $f^d$.
The preserved ``memory'' $h'_{p}$ and feature $f^d$ are concatenated together to perform a linear transformation (convolution), and then activated by a $\tanh$ function.
\begin{equation}\label{Eq:convGRU-ht_bar}
h^{c}_{p}=\tanh { \left( { W }_{ h }\left({ f }^{ d }, { h'_{p} } \right)\right)  } 
\end{equation}
$h^{c}_{p}$ acts similarly to the memory cell in the LSTM and helps the GRF fusion block to remember long term information within the multi-stage fusion.

\noindent
\textbf{Selective Fusion}
$z \odot h_{p}$ :
Indicates how much of the previous features should be preserved.
$(1-z) \odot h^{c}_{p}$: Indicates how much of the current information $h^{c}_{p}$ should be added. Similar to the former, here $(1-z)$ forgets some unimportant information in $h^{c}_{p}$. Or, it can be viewed as a choice of some information in $h^{c}_{p}$.

Combined with $f^{d}$ and ${h}_{p}$, the fusion result at step $p$ is,
\begin{equation}\label{Eq:convGRU-ht}
{ h }_{ q }= z \odot { h }_{ p } + \left( 1 - z \right) \odot { h }^{c}_{p} 
\end{equation}
This operation ignores some information in previous hidden state $h_{p}$, and adds some information from the current step.
Update gate $z$ is equivalent to the \textit{forget gate} in LSTM, and $1-z$ is equivalent to the \textit{input gate} in LSTM. In this way, the \textit{forget gate} $z$ and the \textit{input gate} $(1-z)$ are linked. 
That is, if the previous information is ignored with a weight of $z$, then the information for the current input $h^{c}_{p}$ would be selected with a weight of $(1-z)$. In our case,
if the information in previous stage is depth feature and current input is RGB feature, this enables the complementary information to be effectively merged. Accordingly, the output of the current step ${ f }^{GRF}_{ q }={ h }_{ p }$ will also be passed to the next step.

\noindent
\textbf{Single-stage Fusion Module} The bimodal information passes through multiple DDRs for feature extraction, and each process of the DDR corresponds to a stage. That is, single-stage fusion module has only one layer of DDR from RGB and depth branch, and the fusion block is performed after the DDR. In specific, the GRF module has two input features, $f^d$ and $f^{rgb}$.
At step $1$, the hidden state is initialised as mentioned above. We feed $f^d$ into GRF fusion block, and get the output $h_1$.
Then at step $2$, we reuse the same structure and the same parameters in the GRF fusion block. The input hidden state is replaced by ${h}_{1}$, and the input is the features $f^{rgb}$ extracted from the RGB image.
As shown in Figure~\ref{fig:GRUFusion}, we use the red line with an arrow to indicate the reuse of GRF fusion block at step $2$.

\begin{table*}[t]
\begin{center}
\scalebox{0.91}
{
\begin{tabular} {l |c c c|c c c c c c c c c c c|c} \hline
NYU &  \multicolumn{3}{c|}{scene completion} & \multicolumn{12}{c}{semantic scene completion} \\ 
\hline
method  & prec. & recall & IoU & \cellcolor{rgb1}ceil. & \cellcolor{rgb2}floor & \cellcolor{rgb3}wall & \cellcolor{rgb4}win. & \cellcolor{rgb5}chair & \cellcolor{rgb6}bed & \cellcolor{rgb7}sofa & \cellcolor{rgb8}table & \cellcolor{rgb9}tvs & \cellcolor{rgb10}furn. & \cellcolor{rgb11}objs. & avg. \\ 
\hline
single-stage GRFNet   & 66.5 & {\bfseries85.9} & 60.1 & {\bfseries27.5} & {\bfseries92.9} & 28.1 & 10.7 & 14.9 & {\bfseries60.1} & 33.8 & {\bfseries17.3} & 10.1 & 30.4 & 14.7 & 31.0 \\

multi-stage GRFNet    & {\bfseries68.4} & 85.4 &{\bfseries 61.2} & 24.0 & 91.7 &{\bfseries 33.3} & {\bfseries19.0} & {\bfseries18.1} & 51.9 &{\bfseries 45.5} & 13.4 & {\bfseries13.3} & {\bfseries37.3} & {\bfseries15.0} & {\bfseries32.9}  \\
\hline
\hline
NYUCAD &  \multicolumn{3}{c|}{scene completion} & \multicolumn{12}{c}{semantic scene completion} \\ 
\hline
single-stage GRFNet  & {\bfseries88.4} & 89.1 & 79.7 & 50.0 & 91.4 & 56.4 & {\bfseries18.7} & 41.3 & 56.8 & 52.7 & 33.5 & {\bfseries16.3} & 45.2 & 30.0 & 44.8 \\

multi-stage GRFNet   & 87.2 &{\bfseries 91.0} & {\bfseries80.1} & {\bfseries50.3} & {\bfseries91.8} & {\bfseries58.1} & 18.4 &{\bfseries 42.7} & {\bfseries60.6} & {\bfseries52.8} & {\bfseries34.6} & 11.5 & {\bfseries46.6} & {\bfseries30.8} & {\bfseries45.3} \\
\hline
\end{tabular}
}  
\caption{Results of single-stage GRFNet and multi-stage GRFNet on both NYU and NYUCAD dataset.}
\label{tab:single-vs-multi-stage}
\vspace{-0.5cm}
\end{center}
\end{table*}

\begin{table}[t]
\begin{center}
\scalebox {0.83}
{
\begin{tabular} {l|cc|ccc|c} 
\hline
Method (NYU)	  & GS & MM & prec. & recall & IoU & mIoU \\ \hline
Concatenate Fusion  &                &       & 70.6 & 76.2 & 57.6 & 25.9 \\
Sum Fusion          &                   &       & 67.6 & 79.4 & 57.6 & 25.7 \\
Max Fusion          &                   &       & 67.6 & 79.4 & 57.5 & 25.6 \\
Gated Fusion 	    &      \checkmark    &       & {\bfseries 70.8 } & 77.5 & 58.6 & 27.6 \\ 
LSTM Fusion 	    &     \checkmark    & \checkmark & 68.0 & 82.3 & 59.6 & 28.3 \\
GRF Fusion 	        &     \checkmark    & \checkmark& 66.5 & {\bfseries 85.9} & {\bfseries 60.1} & {\bfseries 31.0} \\
\hline
\hline
Method (NYUCAD) & GS & MM & prec. & recall & IoU & mIoU \\
\hline
Concatenate Fusion  &     & & 87.3 & 83.5 & 74.3 & 37.8 \\ 
Sum Fusion         &   &   & 81.4 & 89.3 & 74.3 & 37.7 \\ 
Max Fusion         &        &  & 81.9 & 87.8 & 73.3 & 36.5 \\ 
Gated Fusion 	    &     \checkmark & & 82.1 & {\bfseries91.3} & 76.0 & 40.2 \\ 
LSTM Fusion 	     &    \checkmark &\checkmark& 83.5 & {\bfseries91.3} & 77.5 & 41.4 \\ 
GRF Fusion 	       &    \checkmark &  \checkmark & {\bfseries88.4} & 89.1 & {\bfseries79.7} & {\bfseries44.8} \\
\hline
\end{tabular}
}
\caption{Results of different single-stage fusion methods on the NYU and NYUCAD dataset. GS denotes Gate Structure, and MM represents Memory Mechanism. With IoU denotes the accuracy of semantic completion and mIoU denotes the accuracy of semantic scene completion.}
\vspace{-0.5cm}
\label{tab:single_fusion}
\end{center}
\end{table}

\begin{figure*}[t]
\centering
{
\includegraphics[width=0.84\linewidth]{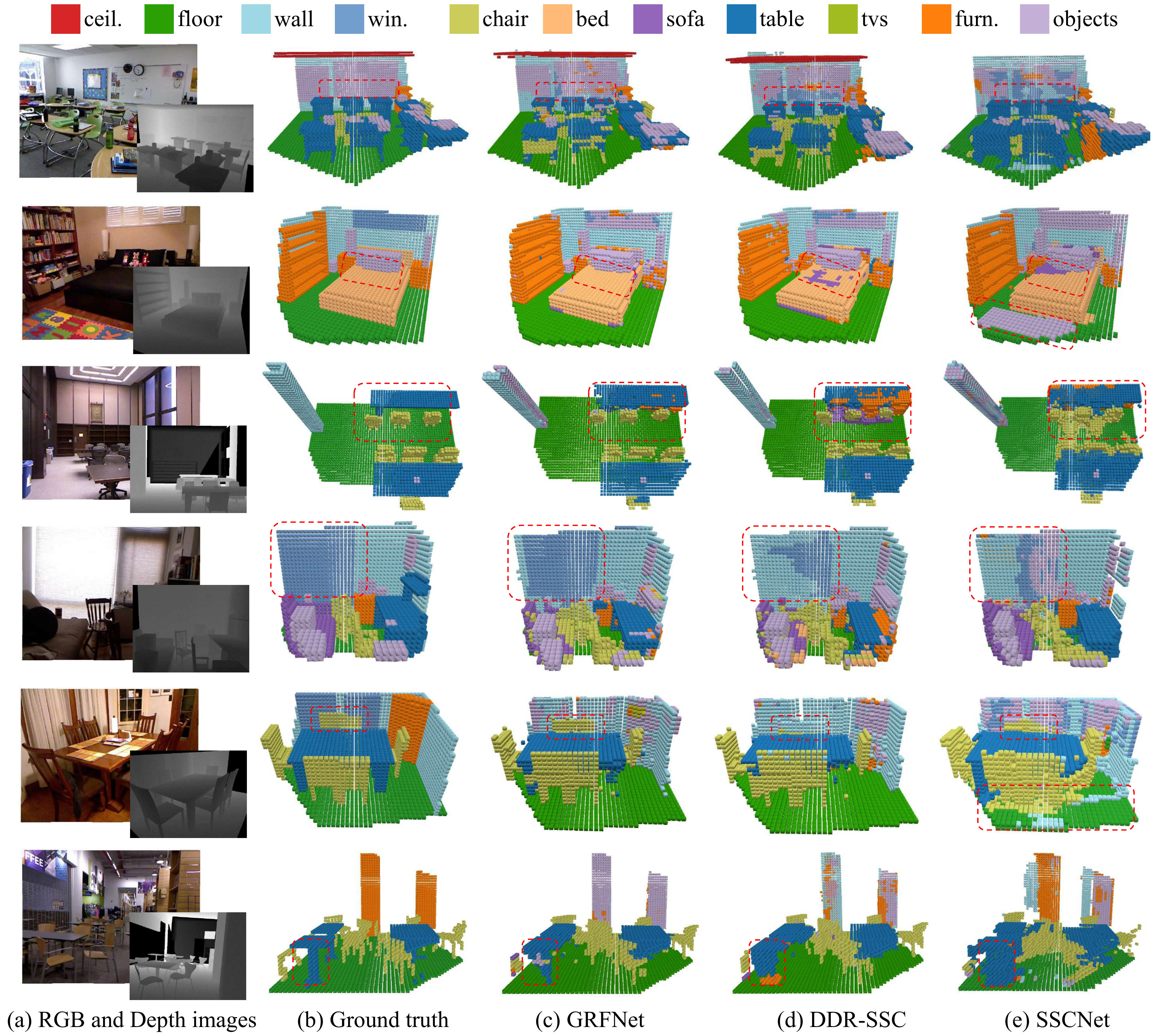}
}
\caption{Qualitative results on NYUCAD. From left to right: Input RGB-D image,
ground truth, results generated by our GRFNet, DDR-SSC~\cite{li2019rgbd}, and SSCNet~\cite{song2017_SSCNet}. 
Overall, our completed semantic 3D scenes are less cluttered and show a higher voxel class accuracy compared to the others. 
}
\vspace{-0.5cm}
\label{fig:viz_GRU}
\end{figure*}

\noindent
\textbf{Multi-stage GRF Fusion Module}
Features extracted by the earlier-stage DDR are at relatively low-level, while those by the later-stage DDR are at relatively high-level regarding to the semantic meaning representation. For multi-stage fusion, the features of the two modal data at each stage will be formed as a sequence.
Taking the $N$-stage RGB-D fusion as an example, the feature sequence is 
$F=\left(f^{d}_{1}, f^{rgb}_{1}, f^{d}_{2}, f^{rgb}_{2}, \cdots, f^{d}_{N}, f^{rgb}_{N}\right)$.
Each feature tensor in $F$ will be fed into the GRF fusion block serially.
The GRU fusion block will be reused $2N$ times and all these fusion stages share the same group of parameters.
Different with the single-stage fusion module which performs the multi-modal feature fusion at only one stage of the network, multi-modal features are fused in multi-stages which covers both of the high-level and low-level features.
It is not only helpful to recover the details of the scene, but also important to propagate information among different stages.

Using the low-level feature to guide high-level feature is a common and reasonable approach in computer vision community. The proposed GRF module can preserves the complementary information and compensates the missing details. Particularly, the color and texture details in RGB image as well as the geometry and distance information in depth are complementary to each other. The ``gate'' structure in the GRF fusion block controls the feature fusion by learning weights (between 0 and 1). Moreover, for multi-stage fusion, the GRF fusion block has the potential to manage the interleaved modalities. To be specific, the bimodal information has gradually transformed into abstract semantic information through the network, and the difference between their distributions is gradually reduced.
Due the above merits, we employ the multi-stage GRF fusion module in our network. The effectiveness of multi-stage fusion module are supported and reflected by our experiments in section~\ref{sec:experiments}.


\vspace{-0.1cm}
\subsection{Training Protocol}
The loss function used in our training process is the softmax cross-entropy loss, and it is performed on the unnormalized network outputs $y$:
\begin{equation}\label{Eq:loss}
\mathcal{L} =-\sum _{ c=0 }^{ N }{ { w }_{ c }{ \hat { y }  }_{ i,c }\log { \left( \frac { { e }^{ { y }_{ ic } } }{ \sum _{ c' }^{ N }{ { e }^{ { y }_{ ic' } } }  }  \right)  }  } 
\end{equation}
where ${ \hat { y }  }_{ i,c }$ are the one-hot ground truth vectors, \ie~${ \hat { y }  }_{ i,c } = 1$ if voxel $i$ is labeled by class $c$, otherwise ${ \hat { y }  }_{ i,c } = 0$. $N$ is the number of classes, and ${ w }_{ c }$ is the loss weight, for balancing different classes, and the setting following SSCNet~\cite{song2017_SSCNet}. 
To compute the loss function, we ignore all voxels outside the field of view but include all voxels inside the view (empty, non-empty and occluded voxels).

We train the network from scratch with the initial learning rate 0.01 which is reduced by a factor of 0.1 after every ten epochs. 
We set the weight of empty voxels ${ w }_{ 0 }$ to $0.05$ for data balancing and increase it by 0.05 for every 40 training epochs. Our model is trained using the SGD optimizer with a momentum of 0.9, weight decay of ${10}^{-4}$ and batch size is 4.

Please note, the order in which the modalities are fed into GRF fusion block has been always fixed (fist Depth, then RGB) for all of the experiments, however, according to our preliminary experiments, using different input order (first RGB, then Depth) has a minor impact on the performance.
\section{Experiments}
\label{sec:experiments}
\subsection{Datasets and Metrics}
\noindent
\textbf{Datasets}
We evaluate the proposed method and compare it with the state-of-the-art methods on NYU~\cite{silberman2012indoor} and NYUCAD~\cite{firman2016NYUCAD} datasets.
NYU consists of 1449 indoor scenes, including 795 training samples and 654 testing samples. The RGBD images of NYU are captured via a Kinect RGBD sensor, and the 3D semantic scene completion labels are from Rock \etal~(\cite{rock2015completing}). The annotations are fitted into the scenes by CAD models.
NYUCAD uses the depth maps generated from the projections of the 3D annotations to reduce the misalignment of depths and the annotations. 

\noindent
\textbf{Metrics}
The primary evaluation metric is the voxel-level intersection over union (IoU) between the predicted labels and ground-truth labels. 
For semantic scene completion, the IoU is calculated for each category on both the observed and occluded voxels. 
For scene completion, all non-empty classes are treated as one category, IoU, precision, and recall of the binary predictions are evaluated on the occupied voxels.

\subsection{Comparisons with the State-of-the-art Methods}
In the task of semantic scene completion, our GRFNet outperforms all existing methods and achieves the start-of-the-art accuracy.
The results on NYU~\cite{silberman2012indoor} and NYUCAD~\cite{firman2016NYUCAD} are shown in Table~\ref{tab:NYU} and Table~\ref{tab:NYUCAD}, respectively.
The GRFNet improves the average IoU over DDR-SSC~\cite{li2019rgbd} by 2.5\% on both NYU and NYUCAD datasets.

The experiments demonstrate that the proposed fusion block effectively utilizes multi-modality information. Since we focus on presenting a practical multi-modal data fusion approach (GRF), we maintain the consistency of the network structure for a fair comparison to prove that the improvement in accuracy comes from the GRF fusion block. In specific, our network framework is the same as DDR-SSC except for the fusion block.

\vspace{-0.1cm}
\subsection{Quantitative Analysis}
\vspace{-0.1cm}
Table~\ref{tab:NYU} shows the quantitative results on NYU dataset~\cite{silberman2012indoor} acquired by our method and other state-of-the-art methods. Approaches of Lin \emph{et al.}~(\cite{lin2013holistic}) and Geiger \emph{et al.}~(\cite{geiger2015joint}) are traditional methods. SSCNet~\cite{song2017_SSCNet}, EsscNet~\cite{zhang2018efficient}, and DDR-SSC~\cite{li2019rgbd} are CNN-based approaches.
Compared to the classical approach SSCNet, the IoUs of GRFNet increase 6.1\% and 8.2\% for SC and SSC tasks, respectively.
In the SSC task, our GRFNet gets 6.2\% higher accuracy than EsscNet and achieves higher IoU in almost every category.
SSCNet and EsscNet only use depth information, while DDR-SSC uses a multi-stage fusion structure to take advantage of the depth and RGB images.
Our GRFNet also uses RGB-D information and achieves a 2.5\% higher average IoU than DDR-SSC.
Regarding the individual class accuracy, the IoUs for each category are also listed out in Table~\ref{tab:NYU}.

As shown in Table~\ref{tab:NYUCAD}, GRFNet achieves outstanding performance on NYUCAD dataset as well. Specifically, compared to Zheng \etal's(~\cite{zheng2013beyond}) and Firman \etal's(~\cite{firman2016NYUCAD}) methods, GRFNet significantly improves the accuracy in two metrics. 
Since SSCNet only employs depth as input, the proposed GRFNet which use RGB and depth information achieves much more accurate results.
Although TS3D~\cite{Garbade2018_twoStream} and DDR-SSC use both RGB and depth information, these methods only adopt simple fusion strategy. On the contrary, GRFNet benefited from the novel fusion block, to obtain 0.7\% and 2.5\% improvements compared to DDR-SSC, and 5.9\% and 3.2\% improvements compared to TS3D for SC and SSC tasks respectively.
In summary, our approach achieves higher accuracy on most indicators than previous methods, especially the average IoU, which reflects the overall performance.
\subsection{Qualitative Analysis}
\label{sec:QualitativeAnalysis}
\vspace{-0.1cm}
Figure~\ref{fig:viz_GRU} visualizes results of the semantic scene completion generated by the proposed GRFNet (c), DDR-SSC (d) and SSCNet (e). We mark the difference in visual quality with a red dotted box for reference.
As can be seen, compared with both SSCNet and DDR-SSC, the scene completion results of our GRFNet are much more abundant in detail and less error-prone. More visualization results and analyses are provided in the supplemental materials.


\subsection{Ablation Study}
\vspace{-0.1cm}
To study the effect of different components and design choices we perform an ablation study. We choose DDR-SSC~\cite{li2019rgbd}
as the baseline, which is the most relevant work to the proposed GRFNet. Since our focus is the fusion strategy, the GRF module will be analyzed in detail below.

\noindent
\textbf{Single-stage Fusion}
To verify the effectiveness of our GRF fusion module, we compare the single-stage GRFNet with a variety of conventional fusion methods, including Concatenate Fusion, Sum Fusion, Max Fusion, and Gated Fusion. 
For better comparison,
we replace the fusion block in the framework (as shown in Figure~\ref{fig:network_structure}) by the compared single-stage fusion methods. The results of the comparison are shown in Table~\ref{tab:single_fusion}.
We have two findings as following: 1) The fusion strategy using the gate structure is better than the one without the gate structure; 2) The memory mechanism can further enhance fusion effects.

As shown in Table~\ref{tab:single_fusion}, Sum Fusion, Max Fusion, and Concatenate Fusion achieve similar performance.
And they are significantly lower than the other three modules in which contain adaptive selection mechanism. LSTM fusion and GRF Fusion have a memory mechanism, but Gated Fusion does not; therefore, the accuracy of the first two methods is better. LSTM is more complex and has more parameters than GRF. However, GRF Fusion is still 2.7\% and 3.4\% more accurate than LSTM on NYU and NYUCAD regarding to SSC accuracy, respectively.

\begin{table}[t]
\begin{center}
\scalebox {0.95}
{
\begin{tabular} {l|ccc|c} 
\hline
Method 				                    & Prec. & Recall & IoU & mIoU \\ \hline
Sum Fusion(DDR-SSC)   & {\bfseries71.5}  & 80.8  & 61.0 & 30.4\\
LSTM Fusion 	                        & 68.0  & 83.2  & 60.2 &  30.2\\   
GRF Fusion 	                            & 68.4  & {\bfseries85.4}  & {\bfseries61.2} & {\bfseries32.9} \\ 
\hline
\end{tabular}
}
\end{center}
\caption{Results of different multi-stage fusion methods on NYU dataset. With IoU represents the accuracy of scene completion, and mIoU denotes the accuracy of semantic scene completion.}
\label{tab:multifusion}
\vspace{-0.4cm}
\end{table}

\noindent
\textbf{Multi-stage Fusion}

1). 
\textbf{Multi-stage Strategy}
In Table~\ref{tab:single-vs-multi-stage}, we compare the performance of single-stage GRFNet and the multi-stage GRFNet. 
Single stage-fusion can only fuse information at one of the network stages.
While multi-stage GRFNet can use both the low-level and high-level information and get higher accuracy than the single-stage version on both datasets.

2).
\textbf{Fusion Strategy}
In DDR-SSC~\cite{li2019rgbd}, sum fusion is used to fuse the four stages of features separately. 
The fusion results are cascaded and handed over to subsequent networks for semantic label prediction.
GRF module employs the recurrent structure that uses the previous fusion results as the input for the next fusion stage, hence the information for each stage can be combined without additional cascading operations.
As can been in Table~\ref{tab:multifusion}, multi-stage GRFNet gets 0.2\% higher average IoU than DDR-SSC on SC and 2.5\% higher on SSC. And GRF fusion is 1\% higher than LSTM fusion on SC and 2.7\% higher on SSC.

\noindent
\textbf{
Parameters and Flops of Different Fusion Stages}
\begin{table}[t]
\begin{center}
\scalebox{1.0}
{
\begin{tabular} {c|c|c} 
\hline

Fusion stages & Params [k] & FLOPs [G] \\ \hline
  1   &   794.59 	&   193.47  \\   
  2   &   803.39	&  	366.65  \\
  3   &   812.19 	&   539.84  \\ 
  4   &   820.99   &    713.02  \\ 
\hline
\end{tabular}
}
\caption{Params and FLOPs of multi-stage GRFNets with different number of fusion stages.}
\vspace{-0.5cm}
\label{tab:ParamsFLOPs-ours}
\end{center}
\end{table}

In Figure~\ref{tab:ParamsFLOPs-ours}, parameters and FLOPs of our network with different fusion stages are listed out. As can be seen, with the increasing of fusion stages, parameters increase slightly, which mainly due to the reuse of GRF fusion block, the only source for more parameters is the new added DDR block (for both Depth and RGB channel). On the contrast, FLOPs increase dramatically, which mainly come from the GRF fusion block and small portion come from DDR blocks. In our implementation, GRF fusion block still employ 3D convolutions, thus bring in relatively high compution costs. As we point out before, our focus of this work is to provide a new strategy for fusing the two-modal data in SSC, which can be improved by light-weight operations.   
\section{Conclusion}
In this paper, we propose GRFNet with a novel gated recurrent fusion module to fuse RGB and depth information.
Different from the existing fusion strategies, we emphasize the importance of the adaptive selectivity of information and the memory mechanism within the fusion block. 
Moreover, we further extend the single-stage GRFNet to a multi-stage version, which can fuse both low-level and high-level feature at different stages.
Our approach has significant advantages over previous methods in multi-modal data fusion and achieves the state-of-the-art performance in semantic scene completion.
Extensive comparison experiments and ablation studies verify the effectiveness of the proposed method. In the future, one of our research interests would be to consider making the proposed GRFNet light-weight, for instance, replacing the 3D convolution of GRF fusion block with DDR. 

{\small
\bibliographystyle{ieee_fullname}
\bibliography{egpaper_for_review}
}


\section{More Details of GRFNet}
\subsection{Detailed Architectures}
The details of the proposed network structure are shown in Table~\ref{tab:networkdetails}. 
PWConv represents the point-wise convolution, and it is used to adjust the number of channels of the feature map.
The down-sample layer in our network is composed of a max-pooling layer and a convolution layer with stride set as 2. The outputs of the two layers are concatenated before fed into the subsequent layers.

\begin{table*}[!htbp]  
\begin{center}
\scalebox{0.99}
{
\begin{tabular}{c|c|c|c|c|c}
\hline
Module                              & Operation         & \begin{tabular}[c]{@{}l@{}}Output Size\\ 2D: $Height \times Width \times Channels$ \\ 3D: $Depth \times Height \times Width \times Channels$ \end{tabular}                       & Kernel             & Stride        & Dilation  \\ \hline
\multirow{8}{*}{Feature Extractor} & PWConv             & $640\times 480\times 8$           & 1         & 1         & 1     \\ \cline{2-6} 
                                    & 2D DDR            & $640\times 480\times 8$           & 3         & 1         & 1     \\ \cline{2-6} 
                                    & 2D DDR            & $640\times 480\times 8$           & 3         & 1         & 1     \\ \cline{2-6} 
                                    & 2D - 3D Projection& $240\times 144\times 240\times 8$ & -         & -         & -     \\ \cline{2-6} 
                                    &  Down-sample      & $120\times 72\times 120\times 16$ & 3         & 2         & 1     \\ \cline{2-6} 
                                    & 3D DDR            & $120\times 72\times 120\times 16$ & 3         & 1         & 1     \\ \cline{2-6} 
                                    & Down-sample       & $60\times 36\times 60\times 64$   & 3         & 2         & 1     \\ \cline{2-6} 
                                    & 3D DDR            & $60\times 36\times 60\times 64$   & 3         & 1         & 1     \\ \hline
\multirow{7}{*}{Feature Fusion}     & GRF stage 1       & $60\times 36\times 60\times 64$   & 3         & 1         & 1     \\ \cline{2-6} 
                                    & 3D DDR            & $60\times 36\times 60\times 64$   & 3         & 1         & 2     \\ \cline{2-6} 
                                    & GRF stage 2       & $60\times 36\times 60\times 64$   & 3         & 1         & 1     \\ \cline{2-6} 
                                    & 3D DDR            & $60\times 36\times 60\times 64$   & 3         & 1         & 3     \\ \cline{2-6} 
                                    & GRF stage 3       & $60\times 36\times 60\times 64$   & 3         & 1         & 1     \\ \cline{2-6} 
                                    & 3D DDR            & $60\times 36\times 60\times 64$   & 3         & 1         & 5     \\ \cline{2-6} 
                                    & GRF stage 4       & $60\times 36\times 60\times 64$   & 3         & 1         & 1     \\ \hline
\multirow{6}{*}{LW-ASPP}            & PWConv            & $60\times 36\times 60\times 64$   & 1         & 1         & 1     \\ \cline{2-6} 
                                    & 3D DDR            & $60\times 36\times 60\times 64$   & 3         & 1         & 3     \\ \cline{2-6} 
                                    & 3D DDR            & $60\times 36\times 60\times 64$   & 3         & 1         & 6     \\ \cline{2-6} 
                                    & 3D DDR            & $60\times 36\times 60\times 64$   & 3         & 1         & 9     \\ \cline{2-6} 
                                    & GlobalAvgPool     & $60\times 36\times 60\times 64$   & -         & -         & -     \\ \cline{2-6} 
                                    & Concatenate       & $60\times 36\times 60\times 320$  & -         & -         & -     \\ \hline
\multirow{4}{*}{Output}             & PWConv            & $60\times 36\times 60\times 160$  & 1         & 1         & 1     \\ \cline{2-6} 
                                    & PWConv            & $60\times 36\times 60\times 12$  & 1         & 1         & 1     \\ \cline{2-6} 
                                    & ArgMax            & $60\times 36\times 60\times 12$   & -         & -         & -     \\ \hline
\end{tabular}
}

\end{center}
\caption{The details of the proposed (GRFNet) network architecture. Including module name, layer operation, output size, kernel
size, stride and dilation.}
\vspace{0.3cm}
\label{tab:networkdetails}
\end{table*}


\subsection{Dimensional Decomposition Residual Block}
In Table~\ref{tab:ConvInDdr3d}, we show the details of the Dimensional Decomposition Residual (DDR)~\cite{li2019rgbd} block. DDR ($k,w,s,d$) denotes the DDR block with the kernel size $k$, the output channels of feature maps $w$, the stride $s$ and the dilation rate $d$. 
DDRConv represents the proposed DDR convolution within a DDR block.

The most significant advantage of using DDR in 3D tasks is the reduction of the number of parameters and the amount of calculation.
In a DDR block, the 3D convolution with the kernel size ${ k }\times { k }\times { k }$ is decomposed into three consecutive layers with filter size $1\times 1\times k$, $1\times k\times 1$ and $k\times 1\times 1$. The most common value for $k$ is $3$.
The computational costs of the original block and the DDR block are proportional to $c^{in}\times c^{out}\times k\times k \times k$ and $c^{in}\times c^{out}\times (k + k + k)$, where $c^{in}$ and $c^{out}$ are the numbers of input and output channels. 

We assume $c^{in} = c^{out} = w$ and ignore bias, then the parameter quantity changes from $w^2\times k^3$ to $w^2 \times 3k$.
The bottleneck structure within the DDR block further reduces its parameter amount and calculation cost.


\subsection{2D to 3D Projection}
Each point in depth can be projected to a position in the 3D space. We voxelize this entire 3D space with meshed grids to obtain a 3D volume. In the projection layer, every feature tensor is projected into the 3D volume at the location corresponding to its position in depth. With the feature projection layer, the 2D feature maps extracted by the 2D CNN are converted to a view-independent 3D feature volume.


\begin{table*}[t]
\begin{center}
\scalebox{0.89}
{
\begin{tabular} {l |c c c|c c c c c c c c c c c|c} \hline
 &  \multicolumn{3}{c|}{scene completion} & \multicolumn{12}{c}{semantic scene completion} \\ 
\hline
method  & prec. & recall & IoU & \cellcolor{rgb1}ceil. & \cellcolor{rgb2}floor & \cellcolor{rgb3}wall & \cellcolor{rgb4}win. & \cellcolor{rgb5}chair & \cellcolor{rgb6}bed & \cellcolor{rgb7}sofa & \cellcolor{rgb8}table & \cellcolor{rgb9}tvs & \cellcolor{rgb10}furn. & \cellcolor{rgb11}objs. & avg. \\ 
\hline

\hline
multi-stage MaxFusion  & 82.5 & {\bfseries 91.8} & 76.8  & 48.2  & 91.4 & 53.3 & 16.5 & 36.5 & 54.6 & 50.5 & 29.3 & {\bfseries 11.8} & 41.2 & 25.0 & 41.7 \\
multi-stage SumFusion & {\bfseries 88.7} & 88.5 & 79.4 & {\bfseries 54.1} & 91.5 & 56.4 & 14.9 & 37.0 & 55.7 & 51.0 & 28.8 & 9.2 & 44.1 & 27.8 & 42.8 \\
multi-stage GRFNet   & 87.2 &91.0 & {\bfseries80.1} & 50.3 & {\bfseries91.8} & {\bfseries58.1} & {\bfseries 18.4} &{\bfseries 42.7} & {\bfseries60.6} & {\bfseries52.8} & {\bfseries34.6} & 11.5 & {\bfseries46.6} & {\bfseries30.8} & {\bfseries45.3} \\
\hline
\end{tabular}
}  
\vspace{0.2cm}
\caption{Results of multi-stage networks with different fusion blocks on NYUCAD dataset. As can be seen, compared with the results based MaxFusion and SumFusion blocks, GRFNet (with GRF fusion block) achieves better accuracy both in scene completion and semantic scene completion tasks. Which may due to the advantage of 'recurrent' property of the proposed GRF fusion block.}
\label{tab:diff_blocks}
\end{center}
\end{table*}



\begin{table}[htbp]
\begin{center}
\scalebox{0.99}
{
\begin{tabular}{c|c|c|c|c}
\hline
Operation & Kernel              & Channels & Stride & Dilation \\ \hline
PWConv    & $1\times 1\times 1$ & $w/4$      & $1$      & $1$        \\
DDRConv   & $1\times 1\times k$ & $w/4$      & $s$    & $d$      \\
DDRConv   & $1\times k\times 1$ & $w/4$      & $s$    & $d$      \\
DDRConv   & $k\times 1\times 1$ & $w/4$      & $s$    & $d$      \\
PWConv    & $1\times 1\times 1$ & $w$        & $1$      & $1$        \\ \hline
\end{tabular}

}
\end{center}
\caption{Details of the DDR ($k,w,s,d$) block. $k$ is the kernel size, $w$ is the output channels of the feature map, $s$ is the stride and $d$ is the dilation rate of the convolution.}
\label{tab:ConvInDdr3d}
\end{table}

\section{Recurrent Property of GRFNet}
\begin{figure*}[t]
\begin{center}
{
\includegraphics[width=0.99\linewidth]{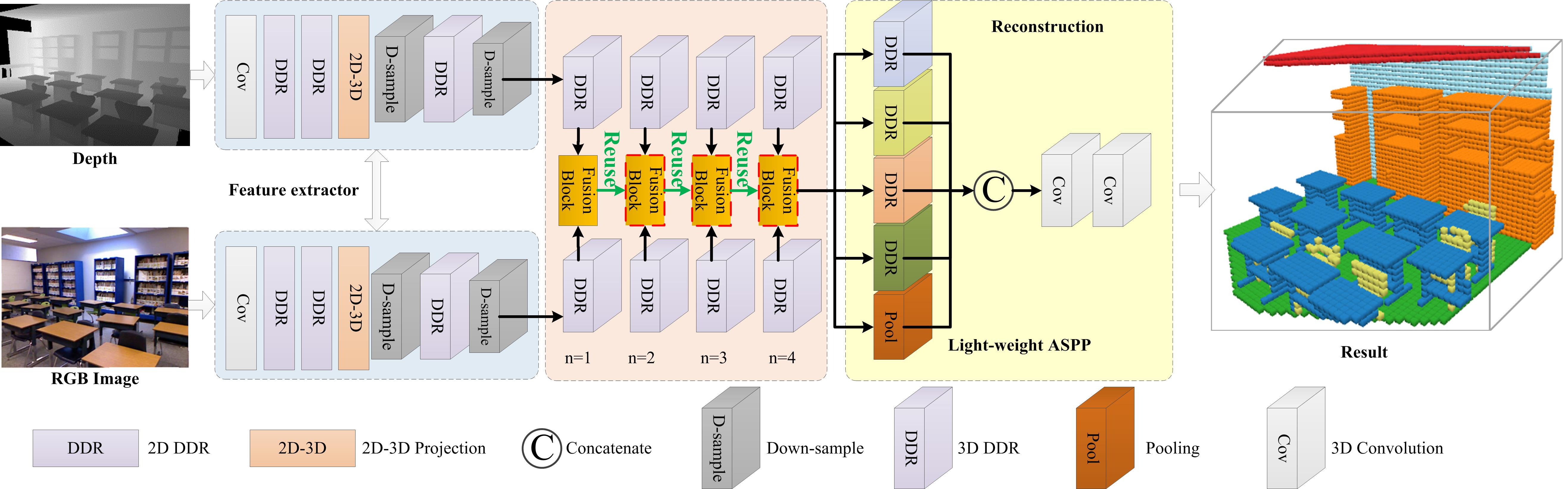}
}
\caption{Network workflow with GRF fusion block}
\label{fig:GRF_fusion}

\end{center}
\end{figure*}

\begin{figure*}[t]
\begin{center}
{
\includegraphics[width=0.99\linewidth]{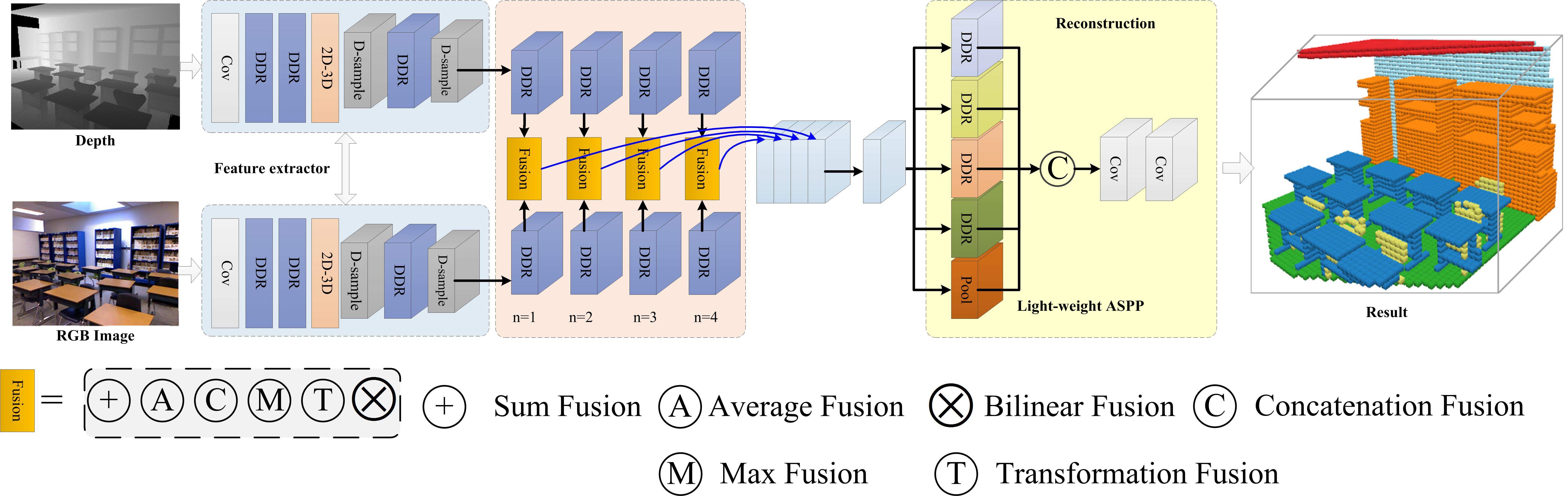}
}
\caption{Network workflow with other fusion blocks}
\vspace{-0.6cm}
\label{fig:other_fusion}

\end{center}
\end{figure*}

In Figure~\ref{fig:GRF_fusion} and Figure~\ref{fig:other_fusion}, network structures with different fusion strategies are listed out. Specifically, Figure~\ref{fig:GRF_fusion} is the network structure with GRF fusion block, which means different stages will share the same GRF fusion block, and it has the `recurrent' property between different stages, that is the low-level fusion could be part of guidance and contribute the following the high-level stages. On the contrary, Figure~\ref{fig:other_fusion} given the network workflow which use other fusion blocks that does not have the 'recurrent' property, including
\textit{Sum Fusion}, \textit{Average Fusion}, \textit{Bilinear Fusion}, \textit{Concatenation Fusion}, \textit{Max Fusion}, \textit{Transformation Fusion}. And as denoted with the blue arrow, different fusion blocks are separated with each other, until they are concated together to be feded to the light-weight ASPP module.

As can be seen in Table~\ref{tab:diff_blocks}, results of networks with different fusion strategies are listed out. The first two rows are the results of networks which employ the MaxFusion and SumFusion, and last row are the results of GRFNet. For both of the scene completion and semantic scene completion tasks, GRFNet boost the performance significantly. We supect that is because the 'recurrent' property of the proposed GRF fusion block.

\section{More Qualitative Results}

\begin{figure*}[t]
\begin{center}
{
\includegraphics[width=0.99\linewidth]{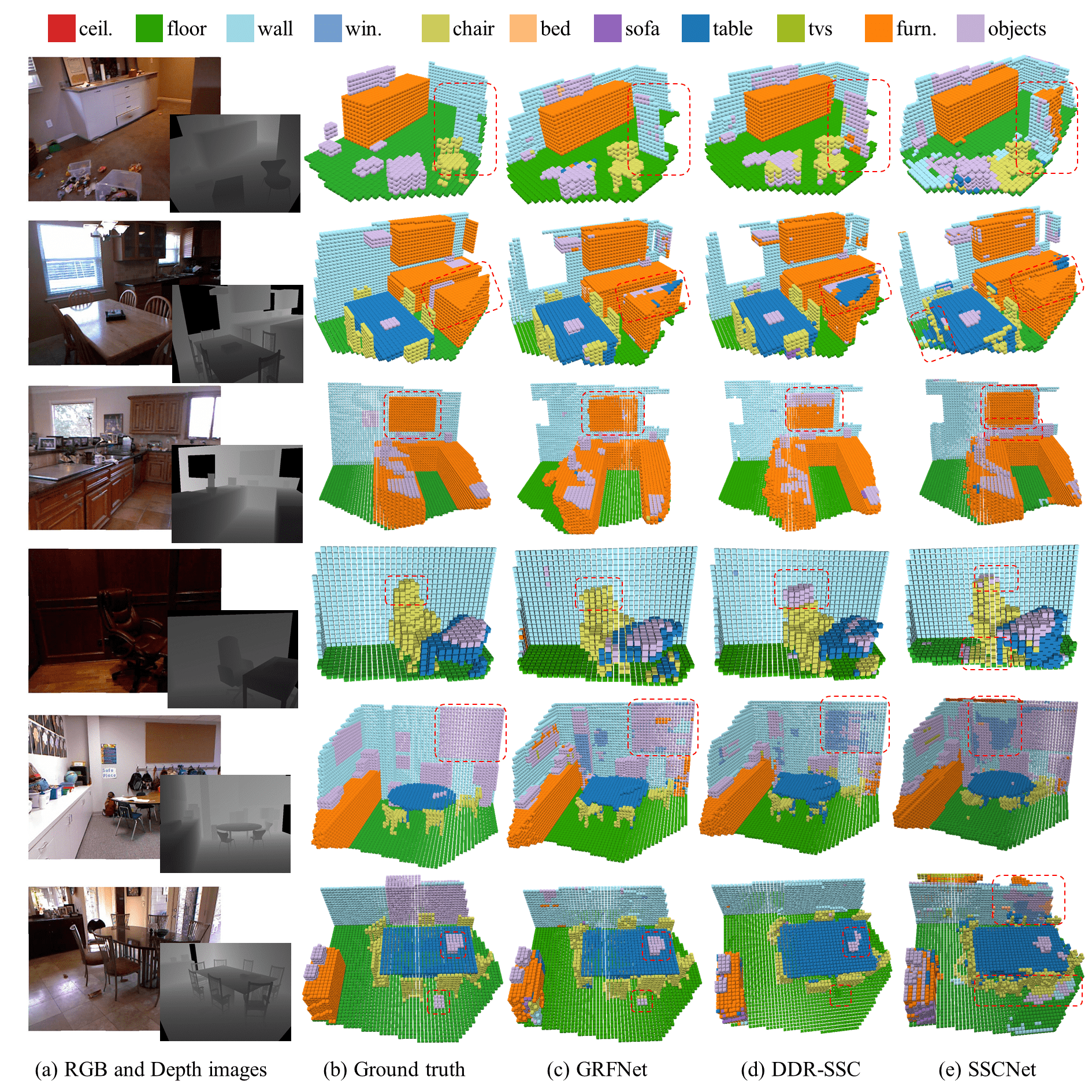}
}
\caption{Qualitative results on NYUCAD. From left to right: Input RGB-D image,
ground truth, results obtained by our proposed GRFNet, results obtained by DDR-SSC~\cite{li2019rgbd} and SSCNet~\cite{song2017_SSCNet}. 
Overall, our completed semantic 3D scenes are less cluttered and show a higher voxel-wise accuracy compared to DDR-SSC and SSCNet.}
\vspace{-0.6cm}
\label{fig:viz_GRF_sup}

\end{center}
\end{figure*}


Figure~\ref{fig:viz_GRF_sup} shows some visualized results on NYUCAD~\cite{firman2016NYUCAD} dataset. As shown in Figure~\ref{fig:viz_GRF_sup}, the proposed GRFNet achieves better results than DDR-SSC~\cite{li2019rgbd} and SSCNet~\cite{song2017_SSCNet}, and is much more accurate in shape completion and semantic segmentation. 
The color information is beneficial for the prediction of semantic labeling.
In Figure~\ref{fig:viz_GRF_sup}, the prediction of furniture in the second, third, and fifth rows is more accurate than the method using only depth.

When an object consists of several parts with various appearances, the RGB information may result in inconsistencies in the semantics of the local areas.
For example, in the first row of Figure ~\ref{fig:viz_GRF_sup}, most of the walls are white, while the right part of the wall is brown. This makes it difficult for the network to predict the semantics of that brown wall.
In this case, the geometric information contained in the depth image can effectively eliminate ambiguity and provide a reasonable inference.

Besides, when different objects are in similar colors, depth can provide adequate information for distinguishing objects.
In the fourth row of Figure ~\ref{fig:viz_GRF_sup}, the chair is very similar to the background in the RGB image, but it can be easily distinguished in depth.
Therefore, in general, the fusion of RGB-D is critical, and our GRFNet can effectively improve the accuracy of SSC.

\end{document}